\documentclass[letterpaper, 10 pt, journal, twoside]{IEEEtran}  

\IEEEoverridecommandlockouts                              

\usepackage{graphics} 
\usepackage{epsfig} 
\usepackage{times} 
\usepackage{amsmath} 
\usepackage{amssymb}  
\usepackage{fixmath}
\usepackage{multirow}
\usepackage[noadjust]{cite}

\usepackage{booktabs} 
\usepackage{threeparttable}  
\usepackage{makecell}   

\usepackage{amsthm}

\usepackage{enumitem}

\usepackage[linesnumbered, ruled, vlined]{algorithm2e}
\usepackage{xcolor}

\SetCommentSty{mycommfont}

\usepackage[font=footnotesize]{subcaption}
\usepackage[font=footnotesize]{caption}

\usepackage[capitalize]{cleveref}
\crefformat{equation}{(#2#1#3)}
\Crefformat{equation}{Equation~(#2#1#3)}
\Crefname{equation}{Equation}{Eqs.}

\usepackage{url}
\usepackage{censor}
\StopCensoring   

\usepackage{xcolor}
\definecolor{r1}{RGB}{0, 0, 0}
\definecolor{r2}{RGB}{0, 0, 0}

\newcommand{\Method}{PinNet}

\author{Yanlong Ma, Nakul S. Joshi, Christa S. Robison, Philip R. Osteen, and Brett T. Lopez%
    \thanks{Manuscript received: May 17, 2026; Revised August 21, 2026; Accepted November 13, 2026.}%
    \thanks{This paper was recommended for publication by Editor Joerg Stueckler upon evaluation of the Associate Editor and Reviewers' comments. This research was sponsored by the DEVCOM Army Research Laboratory (ARL) under SARA CRA W911NF-24-2-0017. Distribution Statement A: Approved for public release; distribution is unlimited.}%
    \thanks{Y. Ma, N. S. Joshi, and B. T. Lopez are with the University of California, Los Angeles, Los Angeles, CA, USA (e-mail: \{yanlong, nakuljoshi, btlopez\}@ucla.edu).}%
    \thanks{C. S. Robison and P. R. Osteen are with the DEVCOM Army Research Laboratory, Adelphi, MD, USA (e-mail: \{christa.s.robison, philip.r.osteen\}.civ@army.mil).}%
    \thanks{Digital Object Identifier (DOI): see top of this page.}%
}

\title{PinNet: Keypoint-Aware Plane-Based Geometric Descriptors \\ for Loop Closure in LiDAR SLAM}

\begin{document}

\maketitle
\setlength{\parskip}{0pt}
\addtolength{\topmargin}{0.00in}  

\begin{abstract}
    Loop closure is essential to reduce drift and build globally-consistent maps in large-scale environments. 
    However, reliable loop closure with only geometric information from, e.g., a LiDAR sensor, remains challenging due to the difficulty of constructing discriminative geometric features.   
    \textcolor{r1}{
    We present \Method{}, a neural network that generates geometry-dependent keypoints and accurate local descriptors for loop closure, including place recognition and point-cloud registration. 
    \Method{} improves descriptor consistency through a keypoint-aware downsampling module that estimates keypoints from high-resolution dense neighborhoods, producing view-consistent local descriptors under different viewpoints.
    It further enhances descriptor discriminability with a plane-based geometric self-attention module that models inter-keypoint spatial relationships using local surface statistics, including Mahalanobis distances and surface-normal angles.
    }
    The approach is comprehensively evaluated on multiple datasets collected with different LiDAR sensors.
    Experimental results demonstrate strong place-recognition performance, precise relative pose estimation, and successful single-shot localization in different environments.
    Video: \protect\url{https://youtu.be/a5auswV1MlY}
\end{abstract}



\section{Introduction}
\label{sec:introduction}

\IEEEPARstart{L}{ocalization} and mapping have attracted significant attention due to the rapid growth of autonomous applications, including self-driving vehicles, infrastructure inspection, and environmental exploration.
Many of these applications require an accurate map that provides a comprehensive understanding of the environment for navigation and decision-making.
Modern LiDAR-based Simultaneous Localization and Mapping (SLAM) algorithms, such as \cite{liosam,fast_lio2,dlio}, have achieved state-of-the-art performance.
However, even with increasingly reliable localization, accumulated drift remains a challenge, particularly in large-scale or feature-sparse environments.
\textcolor{r1}{
Hence, robust loop closure, which combines place recognition and point cloud registration to estimate the relative pose between scans, is essential for providing pose-graph constraints that correct accumulated drift.
}

To this end, prior work has addressed loop closure detection by constructing discriminative descriptors (i.e., features) from LiDAR point clouds \cite{fpfh, m2dp, scan_context}. 
Typically, these approaches rely on handcrafted features and database retrieval to propose candidate loop closures.
Despite recent improvements \cite{ndt-transformer, overlapnet, overlaptransformer, minkloc3d, logg3d, btc} over traditional methods, extracting distinctive geometric features in large-scale real-world environments remains challenging due to point cloud sparsity, viewpoint variations, high computational cost, lack of a global reference frame, and false positives from geometrically similar structures. 
Beyond detecting loop-closure candidates, accurately estimating the relative transformation between point clouds is required to construct a globally consistent map.
However, traditional Iterative Closest Point (ICP) alignment methods are sensitive to initialization and can converge to local minima, reducing accuracy.

\begin{figure}[t!]
  \centering
  \includegraphics[width=0.99\columnwidth]{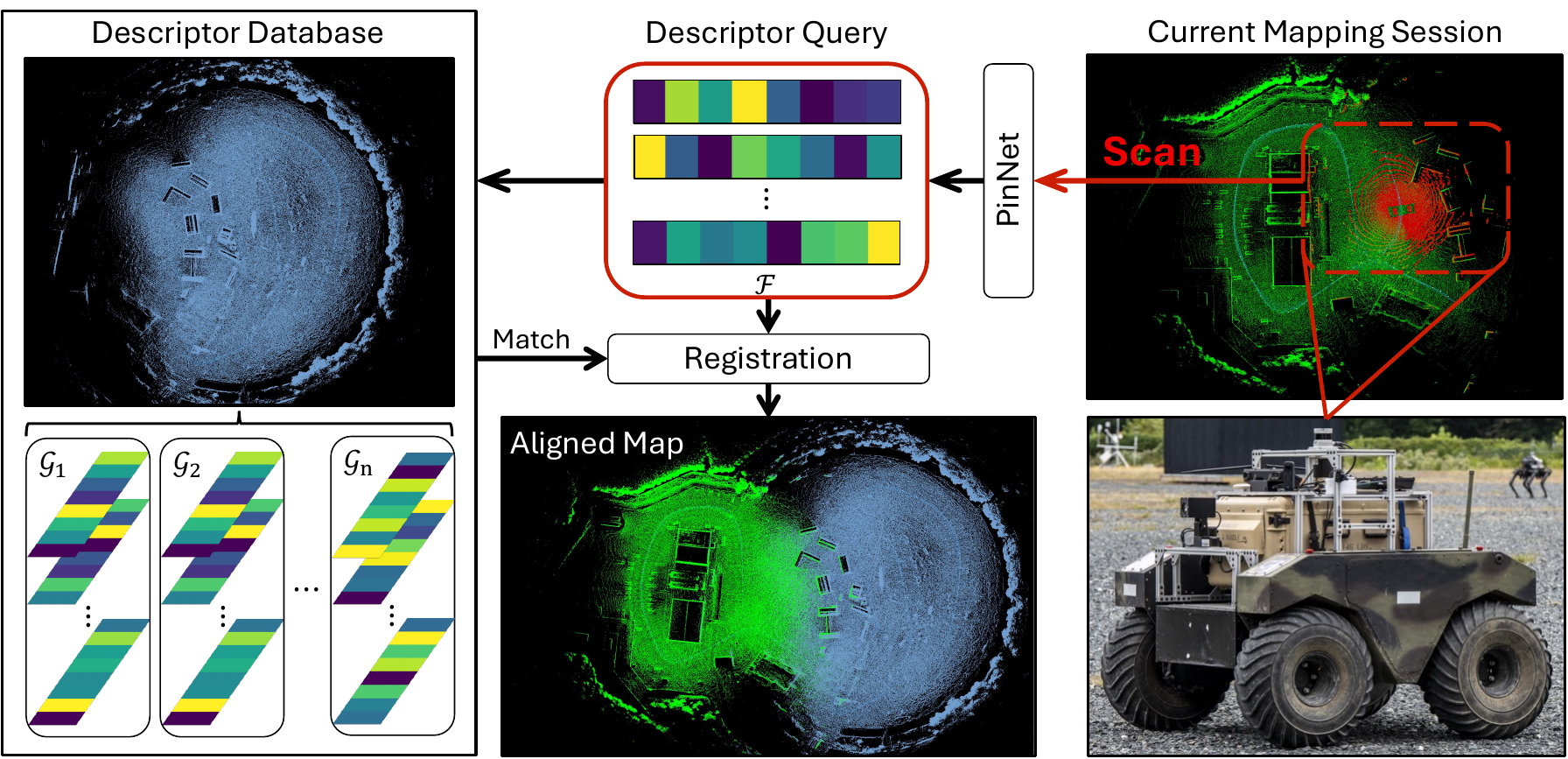}
  \caption{\textbf{\Method{}} provides a robust and accurate place recognition and map alignment method based on learned local geometric descriptors for point clouds.
    A point cloud acquired by the robot used in this work (bottom right) is fed into \Method{}, which generates a set of discriminative local descriptors (top middle).  
    A database of scans from the current or previous mapping session is then queried (left half) for potential matches in descriptor space.
    If a suitable match, i.e., a loop-closure candidate, is found, the two maps are aligned to obtain a globally consistent map (bottom middle).
  }
  \label{fig:cover_fig}
  \vskip -0.08in 
\end{figure}

These challenges highlight the need for geometric descriptors that are discriminative and capable of estimating precise transformations, enabling robust \textcolor{r1}{loop closure detection}.
To \textcolor{r2}{capture more detailed and repeatable local features}, we developed \Method{}, shown in \cref{fig:cover_fig}, a learning-based framework that \textcolor{r2}{combines keypoint-aware local geometric descriptors with plane-based self-attention modules} for loop closure detection. 
By learning distinctive local representations, \Method{} enables reliable loop-closure retrieval and accurate estimation of relative poses between point clouds.

The main contributions of this work are:
\begin{itemize}[itemsep=2pt, topsep=2pt]


\item \textcolor{r1}{A keypoint-aware downsampling module that preserves local geometric details by estimating geometry-dependent keypoints from \textcolor{r2}{high-resolution} dense neighborhoods, producing view-consistent local descriptors.}
\item \textcolor{r1}{A plane-based geometric self-attention module that models \textcolor{r2}{inter-keypoint spatial relationships} through local surface features from nearby dense points, including Mahalanobis distances and surface-normal angles.}
\item \textcolor{r1}{A complete pipeline for both robust place recognition and accurate scan-to-scan registration across diverse environments, different LiDAR sensors, and applications.}
\end{itemize}


We demonstrate the precision and robustness of our approach through a comprehensive quantitative evaluation on the public SemanticKITTI dataset and a qualitative evaluation on self-collected datasets from the \censor{Army Research Lab (ARL) Graces Quarters} test facility and \censor{University of California, Los Angeles (UCLA)} campus, with different types of LiDAR sensors, illustrating the versatility of the designed pipeline and learned descriptors across diverse environments. 
The source code is available at \url{https://github.com/vectr-ucla/pinnet}.

\section{Related Work}
\label{sec:related_work}
Loop closure detection aims to identify previously visited locations to reduce the accumulated drift in simultaneous localization and mapping (SLAM).
This process comprises place recognition, which retrieves potential loop candidates, followed by point cloud registration to estimate the relative transformation and impose pose constraints.
It relies on feature descriptors that provide compact and discriminative representations of point clouds, enabling robust data association despite variations in viewpoint and environmental conditions.

Early works handcraft these descriptors using analytical geometry, surface normals, curvature, or point distributions \cite{fpfh}.
M2DP \cite{m2dp} constructs a global descriptor by projecting point clouds onto multiple 2D planes and computing histogram signatures for each. 
Scan Context \cite{scan_context} and Scan Context++ \cite{scan_context++} construct 2D matrix descriptors by partitioning the point cloud along azimuth and radius, with each bin encoding the height of the points it contains.
However, handcrafted features often lose geometric detail when reducing the dimensionality of point clouds, limiting their accuracy and applicability in large-scale environments.
To mitigate this limitation, other heuristic approaches, such as BTC \cite{btc}, combine a binary descriptor encoding point occupancy along the $z$-axis with a triangle-based descriptor that captures geometric relationships among local point pairs, improving performance.

More recently, learning-based approaches have dominated feature extraction for both place recognition and registration.
PointNetVLAD \cite{pointnetvlad} is an early work in point-based place recognition, combining PointNet \cite{pointnet} for local feature extraction with NetVLAD for aggregation into a single compact global descriptor.
NDT-Transformer \cite{ndt-transformer} represents a 3D point cloud using probabilistic distributions and learns a global descriptor through a transformer-based neural network.
MinkLoc3D \cite{minkloc3d} applies a 3D Feature Pyramid Network to extract local features, while LoGG3D-Net \cite{logg3d} introduces a local consistency loss to improve the local feature extraction. 
LCDNet \cite{lcdnet} designs a shared 3D voxel CNN feature extractor that is applied in both place recognition and registration.
Projection-based methods, such as OverlapNet \cite{overlapnet} and OverlapTransformer \cite{overlaptransformer}, project point clouds into 2D space and use 2D vision techniques for feature extraction. 

Despite these advances, many prior methods omit correspondence estimation or produce unreliable registration results; yet accurate point cloud alignment is essential for imposing loop closure constraints.
Deep Global Registration \cite{deep_global_registration} predicts correspondence confidence with a convolutional network and estimates the transformation via differentiable weighted Procrustes.
\textcolor{r1}{
GeoTransformer \cite{geo-transformer} encodes inter-point relationships using pairwise Euclidean distances and triplet-wise angles from the nearest downsampled points. 
FLAT \cite{flat} further incorporates the Gromov-Wasserstein distance into the geometric embeddings of its cross-attention module. 
Meanwhile, BUFFER-X \cite{buffer-x} targets zero-shot generalization by adaptively selecting voxel size and search radii while generating multi-scale patch descriptors for robustness across diverse scenes.
}
\textcolor{r2}{
Although these methods achieve accurate registration, they assume a corresponding point-cloud pair is given; retrieval therefore requires a forward pass for each query–database pair, resulting in $\mathcal{O}(N^2)$ network evaluations over $N$ scans.
}
\textcolor{r1}{
In contrast, \Method{} extracts distinctive local descriptors with $\mathcal{O}(N)$ evaluations
while maintaining accurate loop-closure retrieval and reducing cost to $\mathcal{O}(m^2d)$ per scan pair, where $m$ is the number of local descriptors per scan and $d$ is their dimensionality, far cheaper than network inference. 
}

\begin{figure*}[t]
\vskip 0.02in
  \centering
  \includegraphics[width=0.99\textwidth]{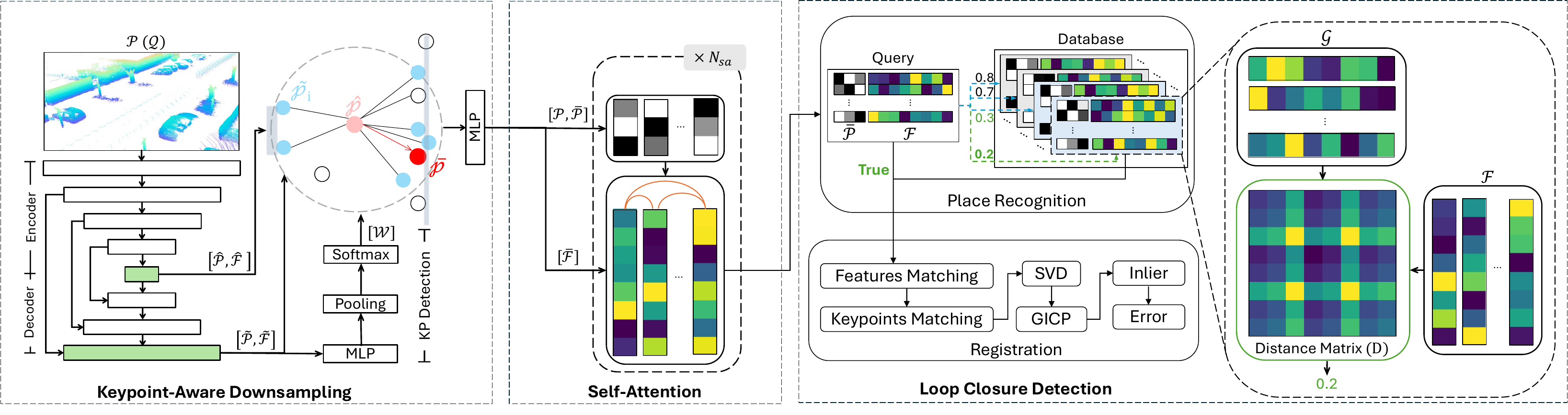}
  \vspace{-1.5mm}
  \caption{The proposed framework consists of three modules: (i) \textit{Keypoint-Aware Downsampling} extracts keypoints and local descriptors from dense point clouds; (ii) \textit{Geometric Self-Attention} enhances the descriptiveness of local feature extraction; and (iii) \textit{Loop Closure Detection} identifies potential loop closures by computing the average of the smallest $s$ pairwise distances between local descriptors, estimates the relative transformation by aligning the corresponding keypoints, and further rejects outliers by evaluating the inlier ratio and relative alignment error.}
  \label{fig:full_sys_arch}
\vskip -0.23in    
\end{figure*}

\section{Methods}
\label{sec:methods}

In this section, we present our proposed framework for geometric feature extraction, enabling both loop closure retrieval (i.e., place recognition) and registration.
The framework consists of three components: (i) a keypoint-aware downsampling module that extracts sparse keypoints and descriptors, (ii) a plane-based geometric self-attention module that enhances the descriptiveness of local features, and (iii) a local-descriptor-based loop closure \textcolor{r1}{retrieval} and registration module.
We first provide an overview of the pipeline, followed by detailed descriptions of each component.

\subsection{System Overview}
\label{subsec:system_overview} 
An overview of the proposed pipeline is shown in \cref{fig:full_sys_arch}. 
It takes dense point clouds as input, either individual point clouds from, e.g., a LiDAR sensor, or keyframes extracted by a keyframe-based SLAM algorithm (e.g., DLIO \cite{dlio,dliom}).
Point clouds are first voxelized to reduce computational complexity.
The voxelized point clouds are then passed to a keypoint-aware downsampling module that extracts keypoints and computes multi-resolution features for keypoints and upsampled dense points.
The self-attention module takes these keypoints and local features as input and applies plane-based geometric self-attention to exploit inter-keypoint spatial relationships.

In the loop closure detection module, loop-closure candidates are identified by thresholding the mean of the smallest local descriptor distances between scans.
The initial relative transformation between the candidate frames is estimated by minimizing point-to-point distances between descriptor-based keypoint correspondences and is further refined using GICP \cite{gicp}.
Finally, false loop-closure candidates are rejected by verifying that both the inlier ratios and the relative pose errors---quantified by the Mahalanobis distance produced by GICP---remain within predefined thresholds.
We now describe each module in our pipeline in more detail.

\subsection{Keypoint-Aware Local Descriptor Generation}
\label{subsec: keypoint_aware_local_descriptor_generation}
Keypoints capture distinctive geometric features of the surrounding 3D structure or scene that are nearly viewpoint-invariant.
This distinctiveness and invariance make keypoints well-suited for robust loop closure detection.
A locally consistent keypoint representation reduces variability in the neighborhood used for discriminative local descriptor generation; furthermore, it decreases the Euclidean error during point correspondence estimation, thereby enhancing the robustness of place recognition and accuracy of the relative transformation estimation between point clouds.

We propose a keypoint-aware local descriptor generation framework that incorporates a keypoint detection module to produce consistent keypoints and local descriptors.
The network structure is illustrated in the Keypoint-Aware Downsampling module of \cref{fig:full_sys_arch}.
To enable real-time pairwise distance computation for local descriptor matching, the raw point clouds are first downsampled to a sparse subset.
We employ the KPConv \cite{kpconv} neural network architecture to hierarchically extract multi-level features for corresponding points during the multi-stage downsampling process.
\textcolor{r1}{
Although many methods build on KPConv, the voxel-grid downsampling strategy used in KPConv introduces inconsistency; even small rotations may alter the distribution of downsampled points and distort the geometry.
}
Furthermore, objects outside overlapping regions can influence the positions of downsampled sparse points within overlapping point clouds, which reduces the consistency of the resulting descriptors.
To address this issue, we introduce a geometry-dependent downsampling strategy that replaces sparse points with so-called keypoints.

\textit{Sparse Point Downsampling and Feature Extraction.} Given a point cloud $\mathcal{P} = \{ p_1, \dots, p_N \, | \, p_i \in \mathbb{R}^3\}$, we apply the KPConv-FPN feature extraction network with fixed voxel-grid downsampling to capture features for both sparse and dense points, doubling the voxel size at each downsampling stage. 
In the encoder's final stage, we denote the sparse points and their features as $\hat{\mathcal{P}} = \{\hat{p}_1, \dots, \hat{p}_n \, | \, \hat{p}_i \in \mathbb{R}^3 \}$ and $\hat{\mathcal{F}} = \{\hat{f}_1, \dots, \hat{f}_{n} \, | \, \hat{f}_i \in \mathbb{R}^{\hat{d}} \}$, where $\hat{d}$ is the sparse-feature dimensionality.
In the decoder, we compute the features of upsampled points up to the first downsampled layer, since the points in $\mathcal{P}$ already fully describe the geometry of the original point cloud without requiring additional descriptors.
We denote the upsampled dense points and their features as $\tilde{\mathcal{P}} = \{\tilde{p}_1, \dots, \tilde{p}_m \, | \, \tilde{p}_i \in \mathbb{R}^3 \}$ and $\tilde{\mathcal{F}} = \{\tilde{f}_1, \dots, \tilde{f}_{m} \, | \, \tilde{f}_i \in \mathbb{R}^{\tilde{d}} \}$, where $\tilde{d}$ is the dense-feature dimensionality.

\textit{Consistent Keypoint Detection.} To address the inconsistencies of the sparse points introduced by voxelization, we estimate new keypoint positions by aggregating the neighboring dense points corresponding to the downsampled sparse points.
Given the sparse points $ \hat{\mathcal{P}}$, we first search for the $k$ nearest neighbors ($k$NN) among the dense points $\tilde{\mathcal{P}}$.
For each sparse point $\hat{p}_i \in \hat{\mathcal{P}}$ with a group of neighboring dense points $\{ \tilde{p}_1^{i}, \dots, \tilde{p}_k^{i} ~ | ~ \tilde{p}_j^{i} \in \tilde{\mathcal{P}} \}$ and their corresponding features $\{ \tilde{f}_1^{i}, \dots, \tilde{f}_k^{i} ~ | ~ \tilde{f}_j^{i} \in \tilde{\mathcal{F}} \}$, we compute the positions of the neighboring dense points relative to each sparse point and calculate the corresponding Euclidean distance offsets.
These relative positions and distances are concatenated with the original point features, yielding $k$ neighboring points augmented with an enhanced feature representation in $ \mathbb{R}^{\tilde{d} + 4}$.

The weights for the $k$ neighboring points, $\mathcal{W} = \{w_1, \dots, w_k \, | \, w_i \in [0,1]\}$, are obtained by feeding the neighboring features into a multilayer perceptron (MLP), followed by a pooling layer, and then normalizing the outputs using a \textcolor{r1}{softmax} function.
The keypoint is computed as a weighted sum of the neighboring points within each group.
The corresponding feature is updated by concatenating the previous sparse point feature $\hat{f}_i$ with the weighted sum of neighboring point features, followed by an MLP.

Additionally, to mitigate overfitting, we employ random dilation grouping as in \cite{rskdd}. 
Rather than selecting the $k$ nearest neighbors of each sparse point, we first identify the $2k$ nearest neighbors and randomly sample $k$ points in each iteration. 
Completing this pipeline, we obtain the keypoints $\bar{\mathcal{P}}$ and their corresponding local descriptors $\bar{\mathcal{F}}$ as a compressed representation of the point cloud.

\subsection{Plane-Based Geometric Transformer Encoder}
\label{subsec:plan_based_geometrical_transformer_encoder}
Although the local descriptors computed as described above accurately encode the local geometry around the keypoints, they do not explicitly capture the global structure of the point cloud.
To address this limitation, we introduce a plane-based geometric transformer encoder to enhance local descriptors with geometric relationships among keypoints.

Inspired by \cite{geo-transformer}, which encodes Euclidean distances and angles between points, we design a plane-based transformer encoder that considers not only point-to-point but also plane-to-plane geometric relationships among keypoints.
The proposed transformer encoder comprises $N_{sa}$ independently configured plane-based geometric self-attention modules. 
Each module employs a conventional self-attention structure augmented with a planar geometric embedding component.
\textcolor{r1}{
Unlike \cite{geo-transformer}, which encodes geometry between keypoints, our geometric encoding leverages local surface features estimated from higher-resolution dense neighborhoods.
}
We consider two metrics to represent planar relationships between keypoints: Mahalanobis distance and surface angle.

\textit{Mahalanobis Distance Embedding.} Given any two keypoints $\bar{p}_i$ and $\bar{p}_j$ in $\bar{\mathcal{P}}$ and the point cloud $\mathcal{P}$, we compute the covariance matrices of the $20$ nearest points to each keypoint, denoted as $\Sigma_i$ and $\Sigma_j$.
The Mahalanobis distance is computed as
$d_{ij} = (\bar{p}_i - \bar{p}_j)^{T} (\Sigma_i + \Sigma_j)^{-1} (\bar{p}_i - \bar{p}_j)$, and the embedding $\varepsilon_{ij}^M \in \mathbb{R}^{\bar{d}}$ is obtained via the sinusoidal function
\begin{equation}
\label{eq:sinusoidal_function}
   \varepsilon_{ij, 2k}^{M} = \sin{\left(\beta_d ~ d_{ij}/c_k \right)} ~~~ \varepsilon_{ij, 2k+1}^{M} = \cos{\left(\beta_d ~ d_{ij}/c_k \right)}
\end{equation}
where $\beta_d$ controls the sensitivity of the Mahalanobis distance and $c_k = (1\mathrm{e}4)^{2k / \bar{d}}$ with dimension of the feature vector $\bar{d}$.

\textit{Surface Angle Embedding.} Given the covariance matrix of a keypoint neighborhood, we take the eigenvector associated with the smallest eigenvalue as the surface normal and denote the normals corresponding to $\Sigma_i$ and $\Sigma_j$ as $n_i$ and $n_j$, respectively. 
The angle between them is computed as
$\alpha_{ij} = \arccos(n_i^\top n_j)$.
The embedding ${\varepsilon}_{ij}^{N}$ can be computed using the same formulation as in \cref{eq:sinusoidal_function}, with the angle $\alpha_{ij}$ and hyperparameter $\beta_n$ replacing $d_{ij}$ and $\beta_d$, respectively.

We also include the point-wise Euclidean distance embedding $\varepsilon_{ij}^{E}$ and the triplet-wise angular embedding $\varepsilon_{ij}^{A}$ proposed in \cite{geo-transformer}.
Finally, the geometric information between keypoints $\bar{p}_i$ and $\bar{p}_j$ is represented as a combination of all embeddings
\begin{equation*}
{\varepsilon}_{ij} = {\varepsilon}_{ij}^{M} {W}^{M} + {\varepsilon}_{ij}^{E} {W}^{E} + {\varepsilon}_{ij}^{N} {W}^{N} + {\varepsilon}_{ij}^{A} {W}^{A},
\end{equation*}
where ${W}^{M}$, ${W}^{E}$, ${W}^{N}$, and ${W}^{A} \in \mathbb{R}^{\bar{d} \times \bar{d}}$ are the respective projection matrices for the Mahalanobis distance, Euclidean distance, surface normal angle, and triplet-wise angular embeddings.
Given the keypoints $\bar{\mathcal{P}}$ and their corresponding features $\bar{\mathcal{F}}$, the new geometric-context-aware local descriptor $f_i$ of keypoint $\bar{p}_i$ is computed as 
\begin{equation*}
    f_i = \sum_{j=1}^{|\bar{\mathcal{P}}|} {\frac{1}{\sqrt{\bar{d}}} \, (\bar{f}_i W^Q) \, (\bar{f}_j W^K + {\varepsilon}_{ij} W^B)^\top \, (\bar{f}_j W^V)},
\end{equation*}
where $W^Q$, $W^K$, $W^V$, and $W^B$ are the projection matrices for the queries, keys, values, and embeddings, respectively.

\subsection{Network Training and Loss Functions} 
\label{subsec:network_traning_and_loss_functions}
We adopt a contrastive learning strategy to train the network.
Given partially overlapping point clouds $\mathcal{P}$ and $\mathcal{Q}$, positive keypoint pairs are defined in overlapping regions, while negative pairs correspond to non-overlapping regions.
The network is trained to minimize descriptor distances for positive pairs and maximize them for negative pairs.
Specifically, for each keypoint, a local patch is constructed by selecting neighboring dense points within a radius. 
Ground-truth correspondences between keypoints are determined by evaluating the average geometric distance between their local patches.
Keypoints are designated as positive pairs if their local patches exhibit at least 10\% overlap; otherwise, they are treated as negative pairs.

We denote the keypoints and local descriptors output by the plane-based geometric transformer as $\bar{\mathcal{P}}$, $\bar{\mathcal{Q}}$, and $\mathcal{F}$, $\mathcal{G}$, respectively.
To train the model to identify accurate correspondences between local descriptors, we adopt the contrastive circle loss, following \cite{circle_loss, geo-transformer}. 
Additionally, since SVD is differentiable, we incorporate a transformation loss $\mathcal{L}_T = \frac{1}{|\mathcal{P}|} \sum_{p_i \in \mathcal{P}} \| (T  - \hat{T}) \, \rho_i \|^2$ to minimize the registration error,
where $\rho_i = (p_i,1)^\top$ and $T,\,\hat{T}\in \mathbb{SE}(3)$ are the ground-truth and predicted transformation between point clouds $\mathcal{P}$ and $\mathcal{Q}$, respectively.
We also use Chamfer loss to supervise keypoint learning by minimizing distances between corresponding keypoints in overlapping regions. 
The overall loss is the equally weighted sum of the three terms.

\subsection{Loop Closure \textcolor{r1}{Retrieval} and Registration}
\label{subsec:loop_closure_detection_and_registration}
The proposed framework represents each point cloud as a set of local descriptors. 
Incorporating the learned keypoints, it enables both loop closure \textcolor{r1}{retrieval} and accurate estimation of relative transformations between point clouds.
Given local descriptor sets $\mathcal{F} = \{ f_1, \dots, f_m \}$ and $\mathcal{G} = \{ g_1, \dots, g_n \}$ extracted from point clouds $\mathcal{P}$ and $\mathcal{Q}$, respectively, we compute the pairwise descriptor distance matrix $D \in \mathbb{R}^{m \times n}$ with entries $D_{ij} = \lVert f_i - g_j \rVert_2$.
The inter-scan distance between $\mathcal{P}$ and $\mathcal{Q}$ is defined as 
$d(\mathcal{P}, \mathcal{Q}) = \frac{1}{s}\sum_{k=1}^{s} \mathfrak{d}_k,$
where $\mathfrak{d}_k$ is the $k^{\mathrm{th}}$ smallest element of $D$ and $s$ is the number of keypoint pairs selected.
Potential loop closures are identified by evaluating whether the inter-scan distance between their corresponding local descriptors falls below a threshold.
When multiple candidates satisfy this criterion, the point cloud with the minimum distance is selected as the candidate.

Let $\bar{\mathcal{P}} = \{ \bar{p}_1, \dots, \bar{p}_m \}$ and $\bar{\mathcal{Q}} = \{ \bar{q}_1,\dots, \bar{q}_n \}$ denote the sets of keypoints extracted from point clouds $\mathcal{P}$ and $\mathcal{Q}$, respectively. 
Following the loop closure \textcolor{r1}{retrieval} scheme, initial correspondences are established by selecting $s$ keypoint pairs with the smallest descriptor distances. 
The coarse transformation $\{R, t\}$ is recovered via SVD. 
This alignment is subsequently refined using 
GICP,
initialized with the SVD solution, to achieve high-precision registration.

Because a single false-positive loop closure can compromise localization and map accuracy, we perform outlier rejection on all loop-closure candidates.
In our pipeline, each candidate is evaluated by first estimating the relative transformation between two point clouds using descriptor-based keypoint correspondences, followed by refinement with dense point cloud alignment. 
Specifically, a candidate is rejected if the GICP alignment error exceeds a predefined threshold or if the post-alignment inlier ratio—the proportion of matching points after alignment—between the aligned point clouds falls below a minimum threshold, indicating a false positive.

\begin{table*}[t]
    \centering
    \caption{Evaluation of Place Recognition on KITTI Dataset}
    \vskip -0.05in
    \label{tab:pr_table}
    \renewcommand{\arraystretch}{1.1}   
    \setlength{\tabcolsep}{4.5pt}       
    \begin{tabular}{l cc cc cc cc cc cc}
        \hline

        \multirow{2}{*}{Sequence} &
        \multicolumn{2}{c}{KITTI 00} &
        \multicolumn{2}{c}{KITTI 02} &
        \multicolumn{2}{c}{KITTI 05} &
        \multicolumn{2}{c}{KITTI 06} &
        \multicolumn{2}{c}{KITTI 07} &
        \multicolumn{2}{c}{Average}\\
        \cmidrule(lr){2-3} \cmidrule(lr){4-5} \cmidrule(lr){6-7} \cmidrule(lr){8-9} \cmidrule(lr){10-11} \cmidrule(lr){12-13}
        & F1 MAX & AP 
        & F1 MAX & AP
        & F1 MAX & AP
        & F1 MAX & AP
        & F1 MAX & AP
        & F1 MAX & AP \\
        \midrule

        BTC~ & 0.9670 & 0.9838 & 0.6418 & 0.5680 & 0.8895 & 0.9623 & 0.9869 & \underline{0.9926} & \underline{0.9106} & \textbf{0.9576} & 0.8792 & 0.8929 \\
        Scan Context~ & 0.9183 & 0.9388 & 0.8151 & 0.7762 & 0.8679 & 0.8699 & 0.9811 & 0.9836 & 0.4319 & 0.3861 & 0.8029 & 0.7909 \\
        OverlapTransformer~ & 0.9309 & 0.9715 & 0.8229 &0.8267 & 0.8553 & 0.9211 & 0.9544 & 0.9818 & 0.5022 & 0.4922 & 0.8131 & 0.8387 \\
        LoGG3D-Net~ & 0.9352 & 0.9733 & 0.8440 & 0.8637 & \textbf{0.9819} & \textbf{0.9918} & 0.9871 & 0.9922 & 0.8291 & 0.9038 & 0.9155 & 0.9450 \\
        LCDNet~ & \underline{0.9693} & \underline{0.9936} & \underline{0.9186} & \underline{0.9599} & \underline{0.9503} & 0.9778 & \underline{0.9945} & \textbf{0.9939} & 0.7757 & 0.8204 & \underline{0.9217} & \underline{0.9491} \\
        \textbf{\Method{} (Ours)}~ & \textbf{0.9813} & \textbf{0.9967} & \textbf{0.9518} & \textbf{0.9849} & 0.9406 & \underline{0.9788} & \textbf{1.0000} & \textbf{0.9939} & \textbf{0.9262} & \underline{0.9461} & \textbf{0.9600} & \textbf{0.9800} \\

        \hline
    \end{tabular}
    \vskip -0.05in  
\end{table*}

\section{Experiments}
\label{sec:experiments}
We quantitatively evaluated the proposed method on the public KITTI dataset and compared our approach with handcrafted and learning-based methods to demonstrate the effectiveness of the learned descriptors for both place recognition and registration.
We qualitatively evaluated loop closure performance through map alignment on datasets collected on the \censor{UCLA} campus and using custom ground vehicles at the \censor{ARL Graces Quarters} test facility (see \cref{fig:cover_fig}).
Moreover, we showed the versatility of the learned descriptors via the task of single-shot localization in a prebuilt point-cloud map.

\subsection{Implementation and Training Details}
\label{subsec:implementation_and_training_details}
We trained the model on KITTI sequences $00$--$09$ using a leave-one-out cross-validation strategy.
Training was conducted on a desktop equipped with an NVIDIA RTX 4090 GPU using the Adam optimizer with an initial learning rate of $5\times10^{-5}$, which was decayed by a factor of $0.95$ every 7 epochs.
The local descriptors $\mathcal{F}$ had a dimension of 256.
We set $s = 256$ when selecting corresponding keypoints and computing pairwise distances to evaluate matching scores and transformations.
All point clouds were voxelized with a leaf size of $v=0.3$~m as the inputs.
For each keypoint, the number of nearest dense points was set to $k = 64$.
The number of self-attention layers $N_{sa}$ was set to 3. 
Embedding hyperparameters $\beta_d$ and $\beta_n$ were set to $1/4.8$ and $1/15^{\circ}$.

To mitigate overfitting, we applied data augmentation during training, including random rotations of up to $\pm 180^{\circ}$ about the yaw axis and $\pm 3^{\circ}$ about the roll and pitch axes.
Random translations of up to $\pm 2$~m, $\pm 2$~m, and $\pm 1$~m were applied along the $x$, $y$, and $z$ axes. 
\textcolor{r2}{These ranges reflect the relative transformations between scans of the same place, with the full yaw range accounting for arbitrary revisit headings.}
Additionally, $10\%$ of the points were randomly dropped, and Gaussian noise $\mathcal{N}(0, 0.01)$ was added for point jittering to simulate occlusions and measurement noise and improve robustness to imperfect LiDAR returns.

\textcolor{r2}{
PinNet contains 34.4 M parameters. 
Processing a point cloud on the same desktop (NVIDIA RTX 4090 GPU) requires 24.1 ms for nearest-neighbor precomputation during downsampling, 43.9 ms for network inference, 6.9 ms for loop-closure retrieval over KITTI 00 with 413 keyframes, and 57.2 ms for registration and outlier rejection (55.22 ms for GICP), for a total of 132.1 ms. 
The running time is substantially shorter than the interval between consecutive keyframes in typical keyframe-based SLAM systems, which enables \Method{} to operate online alongside odometry and mapping.
}

\begin{figure}[t!]
    \centering
    \includegraphics[width=0.97\columnwidth]{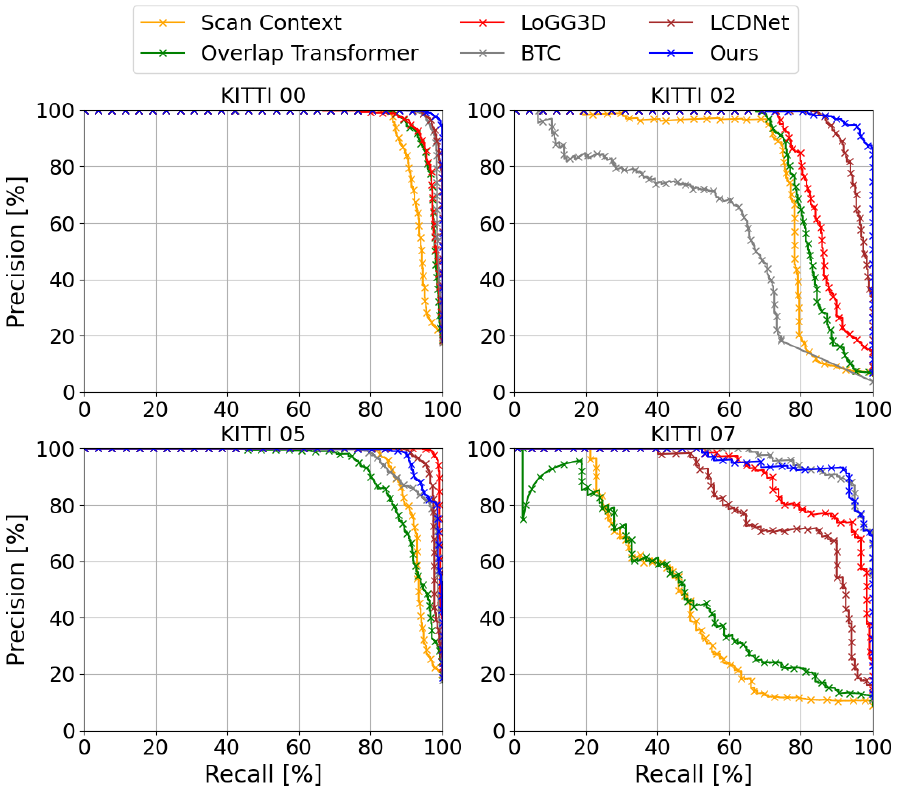}
    \vskip -0.05in
    \caption{\textbf{Precision-Recall curves for KITTI dataset.} The proposed method outperforms the baselines on sequences 00, 02, and 07, while LoGG3D-Net achieves the best performance on sequence 05.}
    \vskip -0.05in
    \label{fig:pr_curves}
\end{figure}
\subsection{Place Recognition}
\label{subsec:place_recognition}
We compared the distinctiveness of the descriptors with several place recognition methods, including the handcrafted methods BTC \cite{btc} and Scan Context \cite{scan_context}, and learning-based methods OverlapTransformer \cite{overlaptransformer}, LoGG3D-Net \cite{logg3d}, and LCDNet \cite{lcdnet}.
We evaluated place recognition on KITTI sequences $00$, $02$, $05$, $06$, and $07$.

For each point cloud, we computed its similarity to all preceding point clouds and selected the one with the minimum descriptor-space distance as the loop-closure candidate.
Precision and recall were computed by varying the similarity threshold.
To avoid temporally adjacent matches, we excluded the most recent 50 point clouds from the search.
A detected loop closure was treated as a true positive if the ground-truth distance was less than $5$~m, and as a false positive otherwise.
Additionally, as in \cite{btc}, we obtained BTC precision and recall by varying the plane-overlap threshold.
For OverlapTransformer, LoGG3D-Net, and LCDNet, we used their released pretrained models for comparison.

\begin{table*}[tb]
    \centering
    \begin{threeparttable}
        \caption{Registration Evaluation on KITTI Sequences}
        \vskip -0.05in
        \label{tab:registration_table}
        \renewcommand{\arraystretch}{1.0}
        \begin{tabular}{l cc ccc ccc ccc}
            \toprule
            \multirow{2}{*}{Approach} &
            \multicolumn{2}{c}{Scan Context} & 
            \multicolumn{3}{c}{LCDNet} &
            \multicolumn{3}{c}{\Method{} w/o GICP}  & 
            \multicolumn{3}{c}{\Method{} w/ GICP} \\
            \cmidrule(lr){2-3} \cmidrule(lr){4-6} \cmidrule(lr){7-9} \cmidrule(lr){10-12}
            & Recall\tnote{†} & Yaw\tnote{†} [deg] & Recall\tnote{‡} & TE [m] & RE\tnote{‡} [deg] & Recall & TE [m] & RE [deg] & Recall & TE [m] & RE [deg] \\
            \midrule
            KITTI 00~ & 95.61\% & 3.255 & 93.12\% & 0.741 & 0.889 & \textbf{100.00\%} & \underline{0.115} & \underline{0.391} & \textbf{100.00\%} & \textbf{0.035} & \textbf{0.155}\\
            KITTI 02~ & 93.66\% & 2.811 & 85.61\% & 1.136 & 1.418 & \underline{97.66\%} & \underline{0.544} & \underline{1.095} & \textbf{99.87\%} & \textbf{0.507} & \textbf{0.993} \\
            KITTI 05~ & 95.43\% & 2.723 & 96.23\% & 0.711 & 0.913 & \underline{98.89\%} & \underline{0.236}  & \underline{0.655} & \textbf{99.07\%} & \textbf{0.136} & \textbf{0.519} \\
            KITTI 06~ & 99.30\% & 0.908 & 75.45\% & 1.359 & 0.794 & \textbf{100.00\%} & \underline{0.091} & \underline{0.268} & \textbf{100.00\%} & \textbf{0.046} & \textbf{0.198} \\
            KITTI 07~ & 87.74\% & 15.14 & 85.18\% & 1.083 & 1.798 & \underline{98.71\%} & \underline{0.250} & \underline{0.469} & \textbf{100.00\%} & \textbf{0.031} & \textbf{0.141} \\
            \bottomrule
        \end{tabular}
        \begin{tablenotes}
            \item[†] As Scan Context estimates only the yaw component, recall is evaluated by checking whether the yaw error is below the rotation threshold.
            \item[‡] The original implementation reported only the yaw error; we provide full rotation error.
        \end{tablenotes}
    \end{threeparttable}
    \vskip -0.15in
\end{table*}

The precision-recall curves are shown in \cref{fig:pr_curves}, with the corresponding \textit{F1-max scores} and \textit{Average Precision (AP)} reported in \cref{tab:pr_table}.
The best method is highlighted in \textbf{bold}, and the second-best is \underline{underlined}.
Our method achieved the highest F1-max score on sequences $00$, $02$, $06$, and $07$, while LoGG3D-Net performed the best on sequence $05$.
Regarding Average Precision (AP), our approach ranked first on sequences $00$, $02$, and $06$, and second on the remaining sequences.
Overall, our method achieved the highest average scores in both F1-max and AP, demonstrating the effectiveness of the proposed descriptors for place recognition.

\subsection{Relative Pose Estimation}
\label{subsec:relative_pose_estimation}
We evaluated the accuracy of relative pose estimation using the learned local descriptors on the same sequences as in \cref{subsec:place_recognition}.
Among the baseline methods, OverlapTransformer and LoGG3D-Net cannot estimate relative poses between point clouds.
Since the released BTC code does not provide the evaluation of pose estimation, we compared our proposed method with LCDNet, which also performs \textcolor{r1}{loop closure detection} within a single network.
Scan Context was included as a baseline.
The RANSAC estimator in LCDNet and the GICP alignment in our method were removed to assess relative pose estimation using only descriptors.


The performance of pose estimation was quantified using all positive loop closures by selecting point cloud pairs whose ground-truth distances were within $5$~m.
Following \cite{lcdnet}, we evaluated pose estimation using three metrics: (1) \textit{Rotation Error (RE)} -- the geodesic distance between rotation matrices, (2) \textit{Translation Error (TE)} -- the Euclidean distance of relative translation, (3) \textit{Recall} -- the ratio of successful registrations where RE $< 5^{\circ}$ and TE $<2$~m.
The results are presented in \cref{tab:registration_table}.
Our method achieved high accuracy in both rotation and translation estimation, outperforming LCDNet and Scan Context across all metrics and sequences.
As a reference, we further assessed performance with GICP refinement, which provides the final estimated transformation in our framework.
With GICP refinement, the average translation error decreased from 0.247~m to 0.151~m, the average rotation error decreased from $0.575^\circ$ to $0.401^\circ$, and the average recall improved from 99.05\% to 99.78\%.

\subsection{Map Alignment with Large Vantage Point Difference}
\label{subsec:map_alignment_exp}
We qualitatively evaluated loop closures generated by the proposed framework in the map alignment task.
\textcolor{r2}{
Note that all loop closure retrieval and registration are performed on individual point clouds rather than on the accumulated map.
}
We used two partially overlapping sequences from \censor{ARL Graces Quarters} to represent the outdoor environment.
We performed loop closure detection on keyframes from \cite{dliom} rather than all frames to reduce the search space.
Specifically, for each keyframe in the query sequence, we detected loop closures against keyframes in the reference sequence.
We then constructed a pose graph from the detected loop closures and optimized it with iSAM2 to align the maps.
As shown in \cref{fig:cover_fig}, the maps align correctly. 
Notably, unlike KITTI, which uses a Velodyne HDL-64E LiDAR, our dataset was collected with an OUSTER OS-1 sensor. 

\begin{figure}[t]
    \centering
    \begin{subfigure}{0.97\columnwidth}
        \centering
        \includegraphics[width=\linewidth]{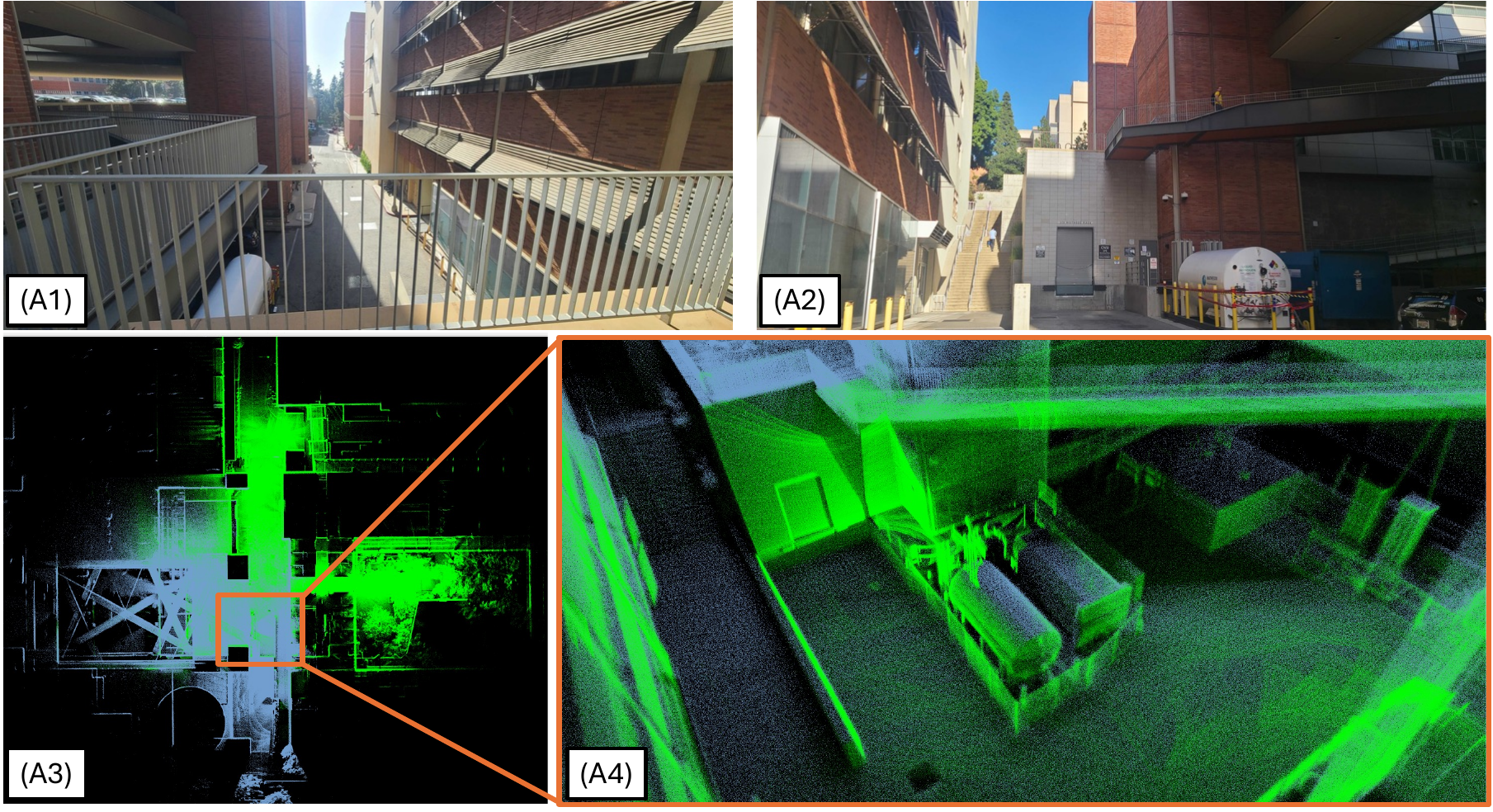}
        \vskip -0.03in
        \caption{A1 shows the image from the upper-level terrace, A2 from ground level, while A3 and A4 show the aligned point cloud.}
    \end{subfigure}
    \begin{subfigure}{0.97\columnwidth}
        \centering
        \includegraphics[width=\linewidth]{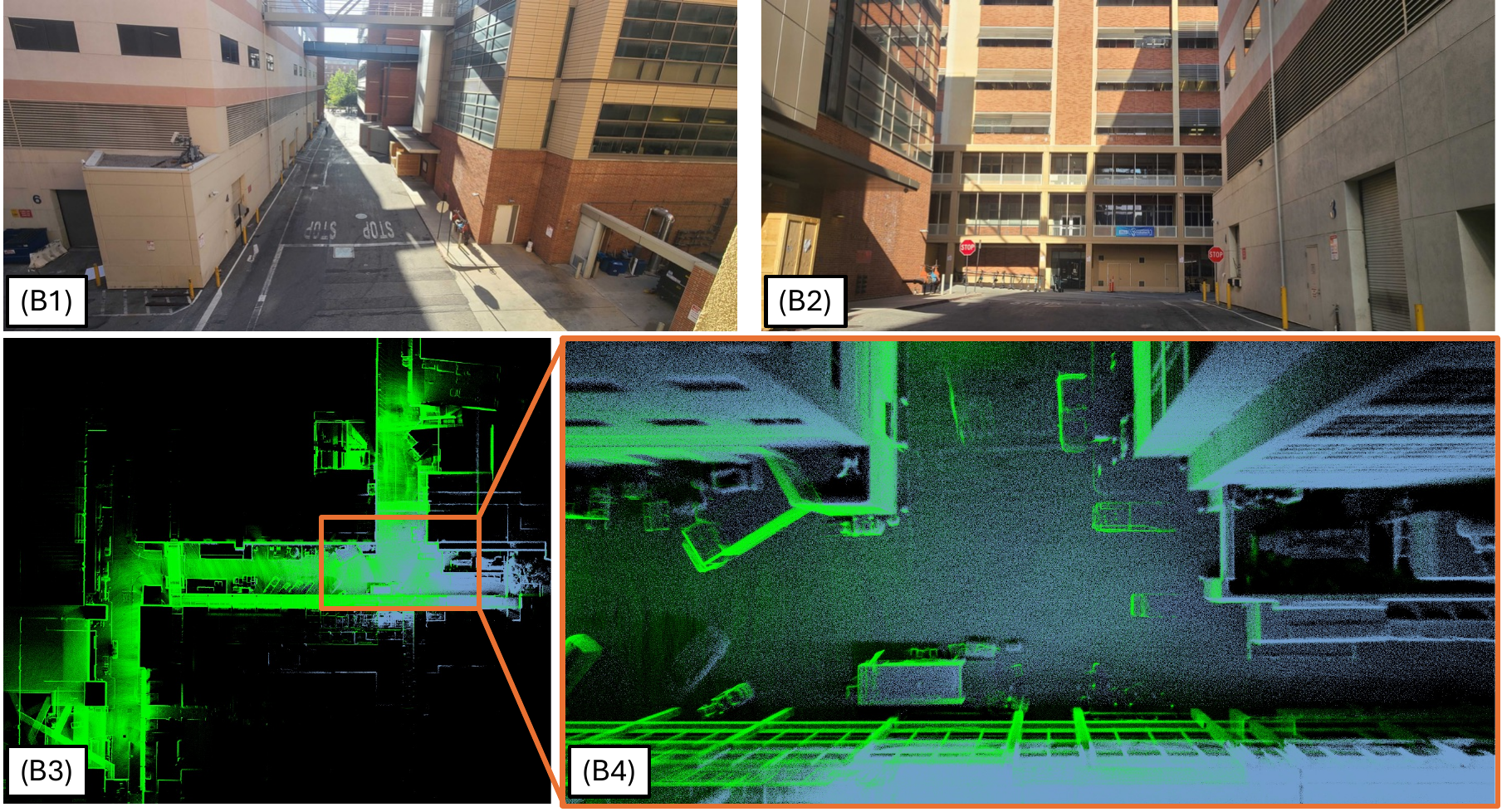}
        \vskip -0.03in
        \caption{B1 shows the image from the outdoor building walkway, B2 from ground level, while B3 and B4 show the aligned point clouds.}
    \end{subfigure}
    \vskip -0.03in
    \caption{Accurate alignment of point clouds from different viewpoints on the \censor{UCLA} campus using a Livox Mid-360 highlights the strong generalization of the proposed local descriptors.
    \textcolor{r2}{
    Loop closures are detected on individual point clouds; accumulated maps are shown for visualization.
    }
    }
    \vskip -0.05in
    \label{fig:ground_aerial_vehicles}
\end{figure}

We further applied the proposed framework to \censor{UCLA} campus datasets collected with a Livox Mid-360 sensor, which exhibit significantly different perspectives.
\textcolor{r2}{
Specifically, the reference dataset was collected from elevated outdoor locations, including upper-level terraces and multi-story walkways, while query scans were acquired at ground level (\cref{fig:ground_aerial_vehicles}), simulating low-overlap alignment between ground and aerial platforms with different sensor mounting configurations.}
For the multi-story walkway datasets, all frames were used due to limited candidates.
Compared to the previous setting, this task is more challenging due to limited overlap and fewer distinctive geometric features along the $z$-axis.
We applied the model without retraining to this dataset, \textcolor{r2}{where the viewpoints of the same place differ in roll and pitch beyond the training range.} 
As shown in \cref{fig:ground_aerial_vehicles}, the maps align accurately, indicating that the proposed descriptors generalize well across sensors, environments, motions, and viewpoints.


\subsection{Single-Shot Localization}
\label{sebsec:single_shot_localization}

We also tested PinNet on single-shot localization to demonstrate the robustness and versatility of the learned descriptors.
Unlike scan-to-scan place recognition followed by pairwise registration, single-shot localization directly estimates the full 6-DoF pose of a query scan within a prebuilt large-scale point-cloud map, making it more challenging due to large viewpoint variations and the need to localize the scan accurately without explicit candidate pairs.

\begin{figure}[tb]
    \centering
    \includegraphics[width=0.99\columnwidth]{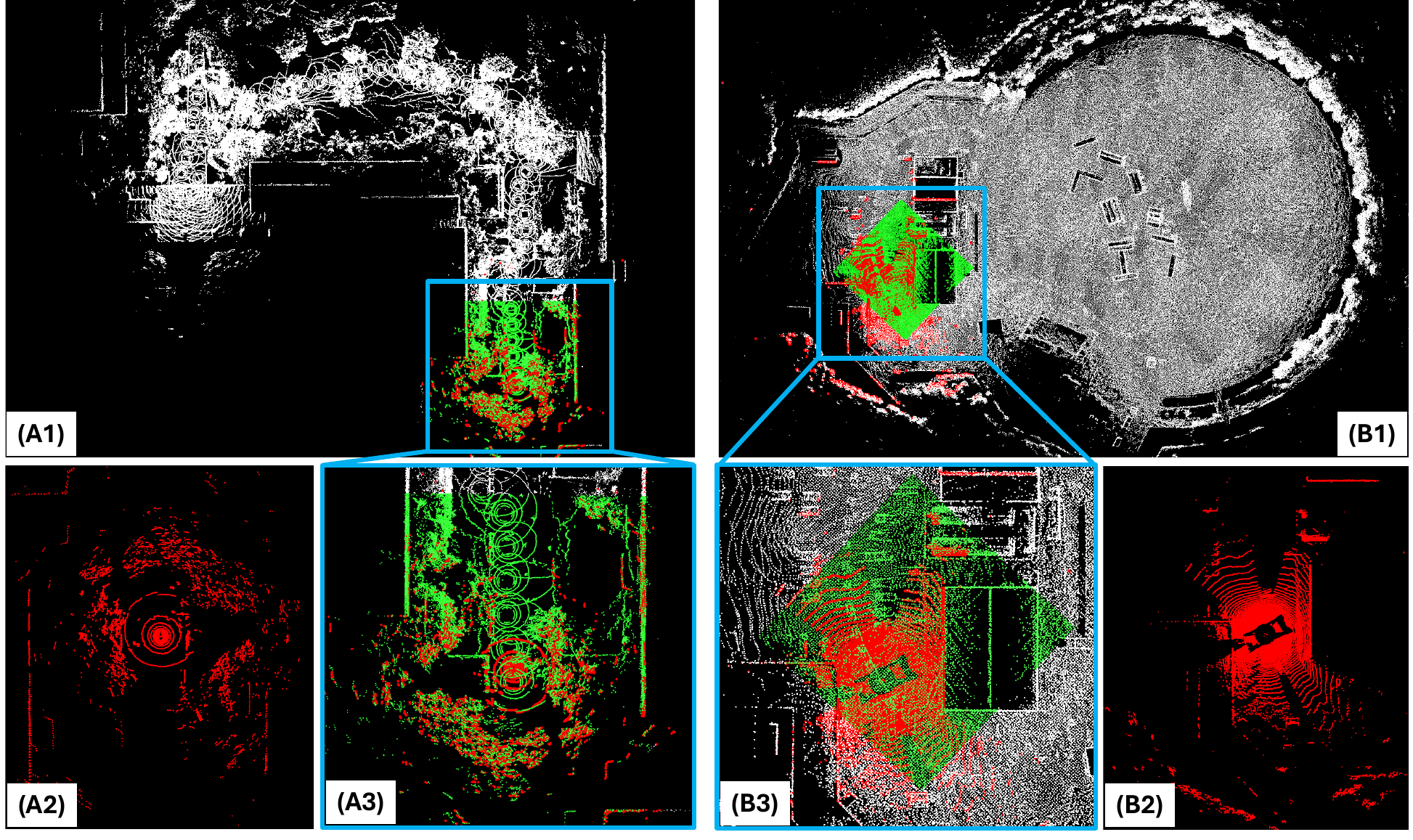}
    \vskip -0.05in
    \caption{\textbf{Single-shot localization.} The global map is visualized in white, the query scans (A2, B2) in \textcolor{red}{red}, and the retrieved partitioned maps (A3, B3) in \textcolor{green}{green}. 
    Query scans are registered to the corresponding regions of the global map, demonstrating accurate localization across diverse environments, including \censor{Bunche Hall (UCLA)} (A) and \censor{Mout Site (ARL)} (B).
    }
    \vspace{-0.05in}
    \label{fig:single_shot_localization}
\end{figure}

To estimate the scan pose, we constructed a descriptor map from a prebuilt point-cloud map by partitioning it into spatially bounded submaps and independently extracting keypoints and local descriptors from each submap. 
This confines the network to local neighborhoods, producing more precise descriptors while avoiding interference from distant structures.
The resulting keypoint-centric descriptors are then aggregated in a common map frame according to their keypoint locations for retrieval and localization.

We retrieved candidate local areas using a sliding-window search over the descriptor map, with a fixed stride across non-empty regions. 
This spatial constraint ensures that matched descriptors remain spatially coherent rather than being scattered across distant regions. 
Formally, given a query scan $p$, the single-shot localization task is formulated as retrieval over the partitioned descriptor maps $\{\mathcal{M}_i\}_{i=1}^{\mathcal{N}}$, with the optimal map index given by
$i^{*} = \mathrm{arg\,min}_{i \in \{1,\dots,\mathcal{N}\}}{d\left( p, \mathcal{M}_i \right)},$
where $d(\cdot, \cdot)$ denotes the inter-scan distance between query $p$ and partitioned map $\mathcal{M}_i$, as defined in 
\cref{subsec:loop_closure_detection_and_registration}.

For robustness, we retained the top three retrieval candidates and selected the one with the highest inlier ratio of dense point correspondences with the query scan.
The query pose was then estimated by solving for the rigid transformation following the registration procedure detailed in \cref{subsec:loop_closure_detection_and_registration}.
If the maximum inlier ratio fell below a predefined threshold, the query scan was classified as non-localizable within the prebuilt map to avoid an incorrect pose estimate.

We validated the proposed framework for single-shot localization on datasets acquired at \censor{UCLA} and \censor{ARL Graces Quarters}. 
Reference maps were generated using the SLAM algorithm \cite{dliom}, while query scans were randomly sampled from separate sequences traversing the same regions to ensure partial overlap.
As illustrated in \cref{fig:single_shot_localization}, red points represent localized query scans, and green points denote the corresponding matched partitioned maps.
The resulting pose estimations are consistent with the ground truth, demonstrating the robustness and versatility of the learned descriptors.


\subsection{Ablation Studies}
\label{subsec:ablation_studies}
\begin{table}[t]
    \centering
    \scriptsize
    \caption{Ablation Studies on KITTI Sequences}
    \vskip -0.05in
    \label{tab:ablation_studies_table}
    \setlength{\tabcolsep}{3.5pt}
    \renewcommand{\arraystretch}{1.0}
    \begin{tabular}{l ccc ccc}
        \toprule
        \multirow{2}{*}{Methods} 
        & \multicolumn{3}{c}{KITTI 05}
        & \multicolumn{3}{c}{KITTI 07} \\
        \cmidrule(lr){2-4} \cmidrule(lr){5-7}
        & \makecell{w/o \\ Keypoints} & \makecell{w/o Plane \\ Embeddings} & \makecell{\Method{} } 
        & \makecell{w/o \\ Keypoints} & \makecell{w/o Plane \\ Embeddings} & \makecell{\Method{} } \\
        
        \midrule
        
        F1 MAX
        & 0.9318 & 0.9247 & \textbf{0.9406}
        & 0.9091 & 0.9076 & \textbf{0.9262} \\
        
        AP 
        & 0.9769 & 0.9762 & \textbf{0.9788}
        & 0.9343 & 0.9374 & \textbf{0.9461} \\
        
        Recall
        & 98.26\% & 98.86\% & \textbf{98.89\%}
        & 88.18\% & 93.65\% & \textbf{98.71\%} \\

        TE
        & 0.393 & 0.270 & \textbf{0.236}
        & 0.755 & 0.523 & \textbf{0.250} \\
        
        RE
        & 0.795 & 0.994 & \textbf{0.655}
        & 1.718 & 0.945 & \textbf{0.469} \\
        
        \bottomrule
    \end{tabular}
    \vskip -0.05 in
\end{table}

\begin{figure}[t!]
    \centering
    \includegraphics[width=\columnwidth]{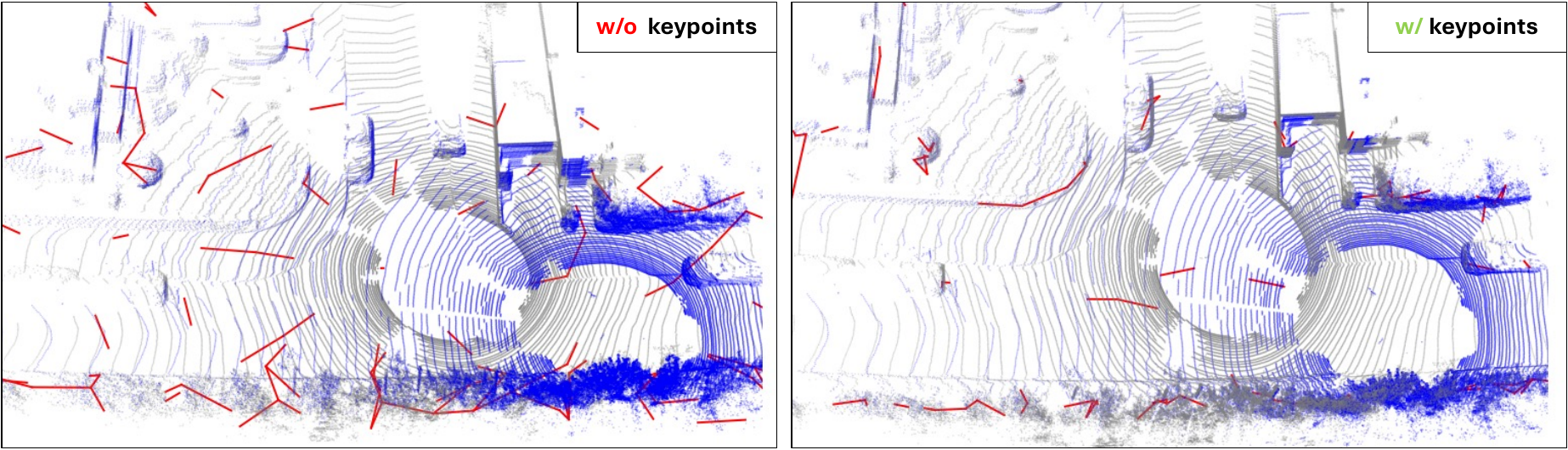}
    \vskip -0.05in
    \caption{
    \textbf{Ablation study of keypoint detection.}
    \textcolor{r2}{
    Lines denote residual vectors between matched point pairs after registration, with lengths proportional to the Euclidean error. 
    Left: without keypoint detection. 
    Right: with keypoint detection. 
    Both panels show the same correspondences; in the right panel, most residuals fall below the rendering threshold and are no longer visible, indicating smaller registration error.
    }
    }
    \label{fig:ablation_residuals}
    \vskip -0.05in
\end{figure}

We assessed the importance of individual components of the proposed framework via ablation studies.
All models were trained and evaluated using the settings in \cref{subsec:implementation_and_training_details}.
We used sequences $05$ and $07$ because they were more challenging in our tests.
The results are reported in \cref{tab:ablation_studies_table}.
Removing the plane-based embeddings reduced the F1-max score by 0.02, 
showing its value for capturing geometric structure for place recognition.
\textcolor{r1}{
While Euclidean distance does not capture local surface geometry, Mahalanobis distance and surface-normal angles capture anisotropic geometry between keypoints.
This distinguishes keypoints with similar local structure but different surface orientations, as reflected in reduced rotation error by $42.03\%$.}
Meanwhile, removing the keypoint detection module resulted in substantially larger transformation error by $0.33$ m, 
demonstrating its importance in reducing drift in point cloud registration. 
The residuals of the matched points, measured as Euclidean distances between the corresponding points used in point cloud registration, are shown in \cref{fig:ablation_residuals}.
A shorter line indicates less variation between matched keypoints, which is clearly reduced when keypoint detection is included.

\textcolor{r1}{
While the keypoint detection module reduces translation errors and improves retrieval performance, we further investigate the effect of the number of neighboring dense points $k$ used to estimate each keypoint.
The results are reported in \cref{tab:ablation_studies_table_kp}.
Larger $k$ improves registration recall and reduces translation error, as keypoints estimated from broader neighborhoods exhibit more spatial consistency. 
However, as $k$ continues to increase, place recognition performance degrades as the keypoints become more concentrated around local centroids, resulting in reduced spatial coverage of the local descriptors.
Conversely, smaller $k$ degrades the registration, as each keypoint is confined to a smaller convex hull, providing less reduction in translation-induced variation and closer to the sparse point it replaces. 
}

\begin{table}[t]
    \centering
    \scriptsize
    \caption{Ablation Studies of Keypoint Neighboring on KITTI Sequences}
    \vskip -0.05in
    \label{tab:ablation_studies_table_kp}
    \setlength{\tabcolsep}{5.5pt}
    \renewcommand{\arraystretch}{1.0}
    \begin{tabular}{l ccc ccc}
        \toprule
        \multirow{2}{*}{} 
        & \multicolumn{3}{c}{KITTI 05}
        & \multicolumn{3}{c}{KITTI 07} \\
        \cmidrule(lr){2-4} \cmidrule(lr){5-7}
        & \makecell{$k=32$}& \makecell{$k=64$} & \makecell{$k=128$} 
        & \makecell{$k=32$}& \makecell{$k=64$} & \makecell{$k=128$} \\
        
        \midrule
        
        F1 MAX
        & \textbf{0.9452} & \underline{0.9406} & 0.9211
        & \underline{0.9180} & \textbf{0.9262} & 0.9076 \\
        
        AP 
        & \textbf{0.9806} & \underline{0.9788} & 0.9721
        & 0.9374 & \textbf{0.9461} & \underline{0.9457} \\
        
        Recall
        & 98.26\% & \underline{98.89\%} & \textbf{99.14\%} 
        & 97.64\% & \underline{98.71\%} & \textbf{99.72\%} \\

        TE
        & 0.283 & \underline{0.236} & \textbf{0.230}
        & 0.262 & \underline{0.250} & \textbf{0.194}  \\
        
        RE
        & \underline{0.739} & \textbf{0.655 }& 0.796
        & \textbf{0.423} & \underline{0.469} & 0.814 \\
        
        \bottomrule
    \end{tabular}
    \vskip -0.05 in
\end{table}

Unlike global descriptors, our local descriptors avoid dependence on a specific aggregation scheme and are less sensitive to partial changes in the point cloud.
To evaluate this design, we retrained the same network with an additional NetVLAD \cite{netvlad} module, a widely used approach for global descriptor generation \cite{lcdnet,overlaptransformer}, and measured point-cloud similarity by the Euclidean distance between the resulting global descriptors.
The global descriptor underperforms the local-descriptor approach across all sequences, with reductions of 0.1264 in F1-max and 0.1141 in AP.


\section{Conclusion}
\label{sec:conclusion}
In this paper, we presented a learning-based loop closure detection framework that achieves accurate and robust performance across diverse environments in both place recognition and point cloud registration tasks.
The proposed keypoint-aware downsampling strategy enhanced the consistency of local descriptors, while the plane-based geometric transformer improved feature extraction by leveraging geometric information from downsampled keypoints. 
We validated the proposed descriptors across place recognition, pairwise registration, map alignment, and single-shot localization, demonstrating their robustness and versatility.
Notably, features extracted from point clouds are not only used for loop closure detection and map alignment but have also been applied to a variety of other tasks in recent studies \cite{optmap}.

\textcolor{r1}{
The main limitation is retrieval cost, which increases with the database size. 
Although comparing descriptors is cheaper than evaluating the network for every scan pair, the retrieval cost scales linearly with database size and may become non-negligible for long sequences, particularly compared with global-descriptor-based retrieval.
Another limitation is that the voxel size and neighborhood radius are tuned for the outdoor spatial scale and may require retuning for environments at different scales.
Finally, while the keypoint module reduces the variation of keypoint positions under $\mathbb{SE}(3)$ transformations, the resulting keypoints are not fully invariant, leading to residual transformation errors.
}

\bibliographystyle{IEEEtran}
\bibliography{ref}


\end{document}